\pdfoutput=1
\documentclass[11pt]{article}

\usepackage{ACL2023}

\usepackage{times}
\usepackage{latexsym}

\usepackage[T1]{fontenc}
\usepackage[utf8]{inputenc}

\usepackage{microtype}

\usepackage{inconsolata}

\usepackage{amsmath}
\usepackage{bm}
\usepackage{balance} 
\usepackage{booktabs}
\usepackage{graphicx}
\usepackage{enumitem}

\usepackage{lipsum}
\usepackage{multirow}
\usepackage{float}

\usepackage{pifont}% http://ctan.org/pkg/pifont
\newcommand{\cmark}{\ding{51}}%
\newcommand{\xmark}{\ding{55}}%

\title{A Multimodal~Dense~Retrieval Approach for Speech-Based Open-Domain Question Answering}

\author{Georgios Sidiropoulos \\
  University of Amsterdam \\
  \texttt{g.sidiropoulos@uva.nl} \\\And
Evangelos Kanoulas\\
  University of Amsterdam \\
  \texttt{e.kanoulas@uva.nl} \\}

\begin{document}
\maketitle
\begin{abstract}
Speech-based open-domain question answering (QA over a large corpus of text passages with spoken questions) has emerged as an important task due to the increasing number of users interacting with QA systems via speech interfaces. Passage retrieval is a key task in speech-based open-domain QA. So far, previous works adopted pipelines consisting of an automatic speech recognition (ASR) model that transcribes the spoken question before feeding it to a dense text retriever. Such pipelines have several limitations. The need for an ASR model limits the applicability to low-resource languages and specialized domains with no annotated speech data. Furthermore, the ASR model propagates its errors to the retriever. In this work, we try to alleviate these limitations by proposing an ASR-free, end-to-end trained multimodal dense retriever that can work directly on spoken questions. Our experimental results showed that, on shorter questions, our retriever is a promising alternative to the \textit{ASR and Retriever} pipeline, achieving better retrieval performance in cases where ASR would have mistranscribed important words in the question or have produced a transcription with a high word error rate.

\end{abstract}

\section{Introduction}
\label{sec:intro}
Voice assistants are convenient, easy to use, and can support users with visual and motor impairments that cannot use conventional text entry devices. Voice assistants such as Amazon Alexa, Apple Siri, and Google Assistant are used daily by everyday users. As a result,  nowadays, users interact with a wide range of commercial Question Answering (QA) systems via such speech interfaces. In other words, millions of users are voicing their questions on virtual voice assistants instead of typing them. That has led to the emergence of speech-based open-domain QA, a QA task on open-domain textual datasets where questions are in speech form.

Most of the research in speech-based open-domain QA focuses on reading comprehension as a component of the commonly adopted \textit{retriever and reader} framework \cite{DBLP:conf/emnlp/FaisalKAA21, DBLP:conf/eacl/RavichanderDRMH21}. However, for effective answer extraction, we need an effective retriever that reduces the search space from millions of passages to the $top$-$k$ most relevant; the performance of passage retrieval bounds that of reading comprehension. Hence, studying passage retrieval in speech-based open-domain QA is crucial. Despite the community attention that passage retrieval for traditional open-domain QA receives \cite{DBLP:journals/corr/abs-2101-00774}, there are surprisingly limited efforts in studying passage retrieval for speech-based open-domain QA.

Recently, \newcite{DBLP:conf/cikm/SidiropoulosVK22} studied passage retrieval for speech-based open-domain QA. 
In particular, following a pipeline approach consisting of an automatic speech recognition (ASR) model for transcribing the spoken question and a dense retriever
\footnote{Dense text retrievers use transformers to encode the question and passage into a single vector each and further use the similarity of these vectors to indicate the relevance between the question and passage.}
for text retrieval (\textit{ASR-Retriever}), they investigated the effectiveness of dense retrievers on questions with ASR mistranscriptions. In their work, they build on the assumption that clean questions are available at training, and the ASR mistranscriptions appear only at inference. That said, such an approach is not directly applicable in a real-world scenario where only the spoken questions are available during training. In a real-world scenario, following an \textit{ASR-Retriever} pipeline would require either (i) using the ASR model to produce the training questions or (ii) training on a different dataset where questions in text form are available.

\textit{ASR-Retriever} pipeline has several limitations. First and foremost, it does not support training in an end-to-end manner. As a direct consequence, ASR propagates its errors to the downstream retriever; the higher the word error rate from ASR, the higher the negative impact on the performance of the retriever. ASR models suffer from mistranscribing long-tail named entities and domain-specific words \cite{DBLP:conf/icassp/HwangMHSGQSSBH22, DBLP:conf/iscslp/MaoKPXHC21, DBLP:conf/interspeech/WangDLLAL20}. The former is of great importance when working on questions because corrupting the main entity of a question can dramatically affect retrieval \cite{DBLP:conf/sigir/SidiropoulosK22}. For example, when the question ``what is the meaning of the name sinead?'' is transcribed as ``what is the meaning of the name chinade?'', then the retrieval will be centered around the wrong entity ``chinade''. Additionally, training an ASR model requires obtaining a large amount of annotated speech which is not always available (e.g., low-resource languages). Finally, during query time, feeding the spoken question to the transformer-based ASR model to obtain its transcription before performing retrieval results in high query latency.

In contrast, we propose a multimodal dual-encoder dense retriever that does not require an ASR model and can be trained end-to-end. Our method adapts the dual-encoder architecture used for dense text retrieval (where both question and passage are in text form) \cite{DBLP:conf/emnlp/KarpukhinOMLWEC20} by replacing the backbone language model used to encode the questions with a self-supervised speech model (Figure \ref{fig:model}). To the best of our knowledge, this is the first multimodal dual-encoder approach for speech-based open-domain QA. Furthermore, in this work, we benchmark passage retrieval for the speech-based open-domain QA using \textit{ASR-Retriever} pipeline approaches, where the spoken questions are transcribed with an ASR model and then used for training the dense text retrievers.

We aim to answer the following research questions:
\textbf{RQ1} How does our ASR-free, end-to-end trained multimodal dense retriever performs compared to its \textit{ASR-Retriever} pipeline counterparts?

\textbf{RQ2} Can our ASR-free retriever alleviate the limitations of \textit{ASR-Retriever} pipelines when it comes to ASR mistranscribing important words in a question?
\textbf{RQ3} How does our ASR-free retriever perform against \textit{ASR-Retriever} when the latter has to deal with previously unseen ASR mistranscription?
\textbf{RQ4} What is the most effective training scheme for learning a multimodal dense retriever?

We make the following contributions:
\begin{itemize}[noitemsep,nolistsep,leftmargin=*]
    \item We benchmark the speech-based open-domain QA task with stronger baselines. In detail, we train the retriever in the \textit{ASR-Retriever} pipeline on spoken question transcriptions and not clean textual questions as in \citet{DBLP:conf/cikm/SidiropoulosVK22}.

    \item We propose an ASR-free, end-to-end trained multimodal dense retriever that can work on speech, and we demonstrate a setup for effective training.

    \item Experimental results showed that our retriever is a promising alternative to \textit{ASR-Retriever} pipelines and competitive on shorter questions.
    
    \item We demonstrate through thorough analysis the robustness of our approach compared to the \textit{ASR-Retriever} pipeline, where its retrieval performance is strongly related to the ASR error; consequently, it varies significantly under different situations. We unveil that pipelines witness a dramatic drop in their retrieval performance (i) as the word error rate of the ASR model increases, (ii) when important words in the spoken question are mistranscribed, and (iii) on mistranscriptions that did not encounter during training. Our ASR-free retriever can alleviate these problems and yield better results. 
\end{itemize}
\section{Methodology}
\subsection{Problem Definition}
\label{sec:problem_def}
% define task
Let $p \in \mathcal{C}$ denote a passage in text form within a passage collection $\mathcal{C}$ in the scale of millions and $q$ a question in speech form. In passage retrieval for speech-based open-domain QA, given $q$ and $\mathcal{C}$, the task is to retrieve a set of relevant passages $\mathcal{P} = \{p_1,p_2, \dots, p_n\}$, where $p_i \in \mathcal{C}$ and can answer $q$.

\subsection{Multimodal Dense Retriever}
\label{sec:method}
In passage retrieval for speech-based open-domain QA, the \textit{ASR-Retriever} pipelines that were used so far suffer from propagating ASR errors to the retriever. Such approaches are not trained end-to-end; thus, the quality of the ASR transcriptions bounds the retrieval performance. Furthermore, the requirement for an ASR model limits the applicability of such pipelines to scenarios where annotated speech is available for training the ASR model. To alleviate the above-mentioned problems, we propose an ASR-free multimodal dense retriever that can support speech, and can be trained end-to-end.
\begin{figure}[t]
    \centering
    \includegraphics[scale=0.65]{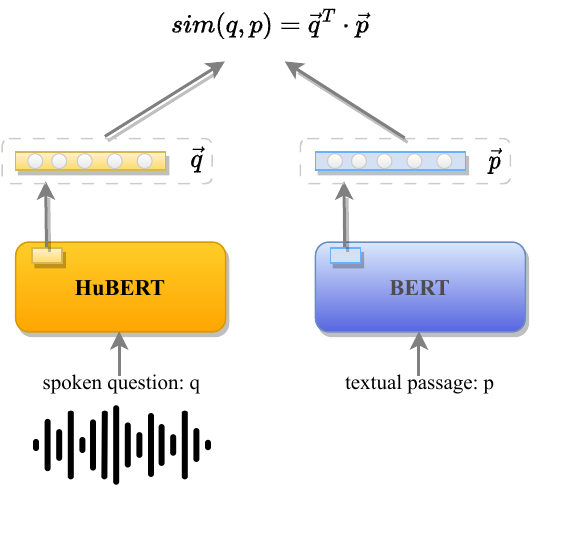}
    \caption{Overview of our multimodal dense retriever.}    
    \label{fig:model}
     % \vspace{-4mm}
\end{figure}

Specifically, we modify the dual-encoder architecture for dense text retrieval in traditional open-domain QA \cite{DBLP:conf/emnlp/KarpukhinOMLWEC20} to account for spoken questions. We replace the BERT-based question encoder with HuBERT \cite{DBLP:journals/taslp/HsuBTLSM21} and leave the BERT-based passage encoder as is. HuBERT is a self-supervised model for speech representation learning, which leverages a BERT-like masked prediction loss. It utilizes offline clustering to provide target labels for masked language model pertaining. Figure \ref{fig:model} illustrates the architecture of our ASR-free, multimodal dense retriever. Suppose a pair of question q, in speech form, and a passage p, in text form. The speech and language encoders produce the output representations:
\begin{equation}
\resizebox{\columnwidth}{!}{$HuBERT(q)=(\bm{q}_1, ..., \bm{q}_n), \mbox{\hspace{5pt}} BERT(p)=(\bm{p}_1, ..., \bm{p}_m)$}.
\end{equation}
We use the first token embedding output from the speech and language modules to encode questions and passages into a single vector each:
\begin{equation}
\vec{q}=\bm{q}_1, \mbox{\hspace{5pt}} \vec{p}=\bm{p}_1.
\end{equation}
The relevance between a question and a passage is computed as the dot product of their vectors:
\begin{equation}
s(q,p) = \vec{q}^T\cdot\vec{p}.
\end{equation}
To this extent, we train our model so that relevant passages to the question (i.e., passages that include the answer) have a higher similarity score than the irrelevant passages. We followed the original dual-encoder training setting from \citet{DBLP:conf/emnlp/KarpukhinOMLWEC20} where, given a question $q$, a relevant passage $p^+$ and a set of irrelevant passages $\{p_1^-, p_2^-, \dots, p_n^-\}$, the model is fine-tuned via the minimization of the softmax cross-entropy:
\begin{equation}
\mathcal{L}_{CE} = -\log \frac{e^{s(q,p^+)}}{e^{s(q,p^+)} + \sum_{p^-}e^{s(q,p^-)}}.
\label{eq:ce}
\end{equation}
The inference phase of our multimodal dense retriever remains the same as in traditional dual-encoders for dense text retrieval. Specifically, we compute the similarity of a question-passage pair as the inner product of the respective question embedding and passage embedding. At query time, only the question needs to be encoded. In detail, we build a dense index of passage vectors (offline) by encoding the whole corpus and storing it in an index structure that supports efficient retrieval of the relevant passages via approximate nearest neighbor search \cite{DBLP:journals/tbd/JohnsonDJ21}. At this point, we want to highlight that we choose a dual-encoder architecture because it has shown high efficiency as a first-stage ranker in large-scale settings. On the contrary, even though cross-encoder architectures can achieve higher performance due to jointly encoding questions and passages, they are not indexable and hence are re-rankers.

\section{Experimental Setup}

\subsection{Baselines and Implementation}
\label{sec:models}
We compare our proposed multimodal method against pipeline approaches that consist of (i) an ASR model for transcribing the spoken question and (ii) a retriever. In detail, similar to \citet{DBLP:conf/cikm/SidiropoulosVK22} we use \textit{wav2vec} $2.0$ \cite{DBLP:conf/nips/BaevskiZMA20} pretrained and fine-tuned on $960$ hours of annotated speech data \cite{DBLP:conf/icassp/PanayotovCPK15} as the ASR model. Concerning the retrievers, we experiment with popular lexical and dense retrievers. We further experiment with dense retrievers explicitly trained to improve robustness against questions with typos since they improve robustness against ASR noise as well \cite{DBLP:conf/cikm/SidiropoulosVK22}.

We need to underline that contrary to \citet{DBLP:conf/cikm/SidiropoulosVK22}, which assumed that the clean textual version of the spoken question is provided for training the retrievers, we follow a real-world scenario where only spoken questions are available; thus, we need to transcribe them for training the retrievers. Our experimental results showed that we build stronger baselines by training on the transcriptions of spoken questions. In particular, we experiment with the following retrievers:

\textbf{BM25} is a traditional retriever based on term-matching. Question and retrieved passages have lexical overlap.

\textbf{Dense Retriever (DR)} is used for scoring question-passage pairs and consists of two separate neural networks (dual-encoder), each representing a question and a passage. Given a question, a positive, and a set of negative passages, the learning task trains the two encoders by minimizing a loss function, typically softmax cross-entropy, to encourage positive question-passage pairs to have smaller distances than the negative ones. To train the \textit{DR} model, we use the training scheme and hyperparameters described in \cite{DBLP:conf/emnlp/KarpukhinOMLWEC20} with a batch size of 64 which is the largest we can fit in our GPU setup.
 
\textbf{Dense retriever with data augmentation (DR+Augm.)} alternates the training scheme of classic dense retrieval via the addition of data augmentation. Recently, there were works that explored a data augmentation approach for robustifying dense retrievers against typoed questions \cite{DBLP:conf/emnlp/ZhuangZ21, DBLP:conf/sigir/SidiropoulosK22}. Specifically, during the training phase of the dense retriever, an unbiased coin is drawn for each question that appears. If tails, the unchanged question is used for training. Otherwise, the question is injected with typos. Typos are sampled uniformly from one of the available typo generators (e.g., random insertion/deletion/substitution, neighbor swapping).

The \textit{DR+Augm.} model we use in our experiments is trained following the training process described in \cite{DBLP:conf/sigir/SidiropoulosK22}. The hyperparameters of the dense retriever remain the same as in the \textit{DR} case.

\textbf{Dense retriever with CharacterBERT and self-teaching (CharacterBERT-DR+ST)} \cite{DBLP:journals/corr/abs-2204-00716} extends \textit{DR+Augm.} by (i) incorporating CharacterBERT, which drops the WordPiece tokenizer and replaces it with a CharacterCNN module, and (ii) by adding a loss that distills knowledge from questions without typos into the questions with typos. The distillation loss forces the retrieval model to generate similar rankings for the original question and its typoed variations. That is achieved by minimizing the KL-divergence:

\begin{eqnarray}
\mathcal{L}_{KL} =  \tilde{s}(q',p) \cdot \log \frac{\tilde{s}(q',p)}{ \tilde{s}(q,p)},
\label{eq:kl}
\end{eqnarray}
where $q$ and $q'$ represent the original question and its typoed variant, respectively while $\tilde{s}$ is the softmax normalized similarity score. The $\mathcal{L}_{KL}$ loss (Equation \ref{eq:kl}) is combined with the supervise softmax cross-entropy loss $\mathcal{L}_{CE}$ to form the final loss:
\begin{eqnarray}
\mathcal{L}_{ST}=\mathcal{L}_{KL}+\mathcal{L}_{CE}.
\end{eqnarray}

For training \textit{CharacterBERT-DR+ST}, we follow the original work by \citet{DBLP:journals/corr/abs-2204-00716}.
 
\subsection{Datasets and Evaluation}
\label{sec:dataset}
We conduct our experiments on the spoken versions of MSMARCO passage ranking \cite{DBLP:conf/nips/NguyenRSGTMD16}, and Natural Questions \cite{DBLP:journals/tacl/KwiatkowskiPRCP19} introduced by \citet{DBLP:conf/cikm/SidiropoulosVK22}. In the Spoken-MSMARCO and Spoken-NQ, the question is in spoken form, while the passage is in the form of text. For the former dataset, the underlying corpus consists of $8.8$ million passages, $~400K$ training samples, and  $6,980$ development samples, with an average sample duration of $3sec$ (${\sim}6$ words). Spoken-NQ facilitates $58,880$ training, $6,515$ development, and $3,610$ test samples with an average sample duration of $3.86sec$ (${\sim}9$ words). Questions can be answered over Wikipedia. Concerning the pipeline approaches that employ ASR for the transcription of the spoken questions, the WER for Spoken-NQ (test) is $20.10\%$ and for Spoken-MSMARCO (dev) is $32.87\%$.

To measure the retrieval effectiveness of the models on Spoken-MSMARCO, we use the official metric of the original MSMARCO, namely, Mean Reciprocal Rank ($MRR@10$) and the commonly reported Recall ($R@50, R@1000$). Similar to the original MSMARCO, we report the metrics on the development set, since the ground-truths for the test set are not publicly available. Following previous works on NQ and Spoken-NQ, we report we use answer recall at the $top$-$k$ ($AR@20, AR@100$) retrieved passages. Answer recall measures whether at least one of the $top$-$k$ retrieved passages contains the ground-truth answer string, then $AR@k=1$ otherwise $AR@k=0$.

\subsection{Implementation Details}
\label{sec:details}
We train our multimodal dense retriever using the in-batch negative setting described in \cite{DBLP:conf/emnlp/KarpukhinOMLWEC20}; with one hard negative passage per question and a batch size of 64, the largest batch we could fit in our GPU setup. The question and the passage encoders are implemented by \textit{HuBERT-Base} \cite{DBLP:journals/taslp/HsuBTLSM21} and \textit{BERT-Base} \cite{DBLP:conf/naacl/DevlinCLT19} networks, respectively. We take the first embedding from the two output representations of each sequence to represent their corresponding speech and text sequences. Additionally, our experimental results showed that having different learning rates for the question and passage encoders leads to more effective training. Specifically, we set the learning rate to $2e$-$4$ for the question encoder and $2e$-$5$ for the passage encoder. To this end, we train with the Adam optimizer and linear scheduling with $0.1$ warm-up for up to 80 epochs for the small Spoken-NQ dataset and 10 for the larger Spoken-MSMARCO dataset. 

To end up with the abovementioned parameters, we searched learning rates ranging from $2e$-$6$ to $2e$-$4$ (for cases where question and passage encoders share the same learning rate or have different ones) and explored \textit{HuBERT-Base} and \textit{Wav2Vec2-Base} question encoders. We also experiment with warming up the question embedding space before training our retriever, following the sequence-level alignment approach described in \citet{DBLP:conf/naacl/ChungZZ21}. However, we did not see any improvements. We chose the best hyperparameters with respect to the best average rank in the development split \cite{DBLP:conf/emnlp/KarpukhinOMLWEC20}.

\section{Results \& Discussion}

\subsection{Main Results}
\label{sec:main_results}
\begin{table*}[ht]
\resizebox{\textwidth}{!}{%
\begin{tabular}{@{}l|l|c|cc|ccc@{}}
\toprule
\multirow{2}{*}{} &
  \multirow{2}{*}{Training questions} &
  \multirow{2}{*}{End-to-end} &
  \multicolumn{2}{c}{Spoken-NQ (test)} &
  \multicolumn{3}{c}{Spoken-MSMARCO (dev)} \\
            &     &             & AR@20  & AR@100  & R@50 & R@1000       & MRR@10      \\ \midrule
BM25        & -    & \xmark   & 36.98  & 52.49   & 24.71 & 45.34        & 6.97        \\
DR & Transcriptions    & \xmark  & 68.36  & 78.53   & 52.06 & 79.35        & 17.74       \\ 
DR+Augm.   &  Transcriptions \& Typos   & \xmark  & \textbf{69.77} & 80.05 & 52.97 & 80.56  & 17.87       \\
CharacterBERT-DR+ST & Transcriptions \& Typos & \xmark & 68.22 & \textbf{80.3} & \textbf{53.88} & 78.95 & \textbf{20.53} \\ \midrule
Multimodal DR (Ours)&  Speech   & \cmark & 57.25       & 70.11        & 51.37 & \textbf{81.34}         & 15.77       \\ \bottomrule
\end{tabular}%
}
\caption{Retrieval results on Spoken-NQ and Spoken-MSMARCO. End-to-end ``\xmark'' accounts for \textit{ASR-Retriever} pipeline approaches that can not be trained in an end-to-end manner.}
\label{tab:multimodal_vs_pipeline}
\end{table*}
To answer \textbf{RQ1}, we compare the retrieval performance of our multimodal dense retriever against the \textit{ASR-Retriever} pipelines we described in Section \ref{sec:models}. From Table \ref{tab:multimodal_vs_pipeline} we note that our model is highly competitive on the Spoken-MSMARCO dataset, while the pipeline approaches perform significantly better on the Spoken-NQ dataset. This discrepancy is twofold. First, our multimodal dense retriever performs better on shorter questions. We conjecture that the low performance of our model on longer questions is due to encoding the spoken question into a single vector which might not be enough to capture the necessary information as the length of the question increases. Second, the higher the word error rate the higher the negative impact on the \textit{ASR-Retriever} pipelines. Spoken-MSMARCO has shorter questions and higher word error rate compared to Spoken-NQ (Section \ref{sec:dataset}).

\begin{figure}[!h]
        \centering
        \includegraphics[width=1\linewidth]{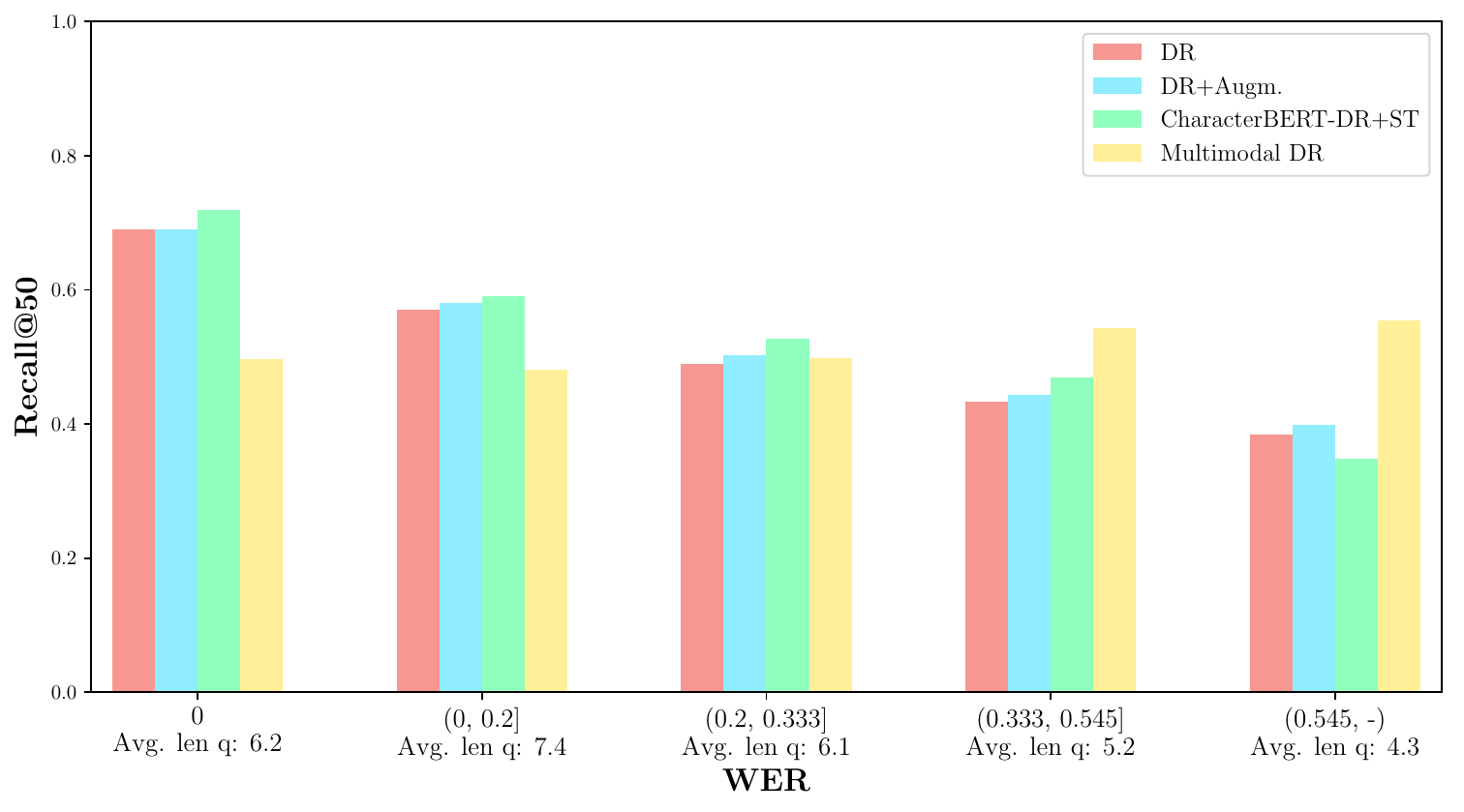}
        \caption{Retrieval performance w.r.t the WER of questions; on Spoken-MSMARCO (dev). Each bin with a non-zero WER has ${\sim}1300$ samples, while the one with a zero WER has ${\sim}1600$ samples. We also report the average question length, in tokens, per bin.}
        \label{fig:wer}
\end{figure}

In Figure \ref{fig:wer}, we verify our aforementioned claims. Figure \ref{fig:wer} reports the retrieval performance of the \textit{ASR-Retriever} pipelines concerning the different word error rates of the ASR model and where our ASR-free method stands. 
Pipelines show strong results when no corrupted words are in the transcriptions (WER is $0$). However, there is a significant drop in the retrieval performance of all the pipeline approaches as the word error rate increases. 
On the other hand, our ASR-free, multimodal dense retriever is significantly more stable across the different settings. At the same time, we see an increase in the performance of our approach as the length of the question decreases (see the average length under each bin in Figure \ref{fig:wer}).
That said, we can conclude that adopting our multimodal dense retriever as the word error rate of the ASR model increases yields better results.

Despite the strong performance that \textit{ASR-Retriever} pipelines achieve, there are several limitations we need to highlight. In Table \ref{tab:params_data_time}, we compare our model to \textit{ASR-Retriever} pipelines in terms of query time, the need for annotated speech data, and parameters. A major constraint for pipelines is the requirement in annotated speech for training an ASR model. In the real world, such data are not always  in abundance. Annotated speech can be hard to obtain when dealing with low-resource languages or specialized domains such as the medical domain, where a general-purpose ASR system will underperform. In such scenarios, the applicability of \textit{ASR-Retriever} pipelines is limited. In contrast, our approach is ASR-free,  hence, does not need annotated speech. Regarding query time, as shown in Table \ref{tab:params_data_time}, the query time of our model is substantially shorter than that of the \textit{ASR-Retriever}. Passing the spoken question through an ASR model to obtain its transcription introduces additional overhead. 

In our work, we are interested in the cases where ASR generates transcriptions with a higher word error rate; therefore, we conduct extensive analysis focusing on such cases.
\begin{table}[!h]
\resizebox{\columnwidth}{!}{%
\begin{tabular}{@{}lllll@{}}
\toprule
 & \begin{tabular}[c]{@{}l@{}}Annotated\\ Speech\end{tabular} & \begin{tabular}[c]{@{}l@{}}ASR\\ $\#$params\end{tabular} & \begin{tabular}[c]{@{}l@{}}Retriever\\ $\#$params\end{tabular} & Time \\ \midrule
DR+Augm. & 960h & 95M & 220M & 45.3ms \\
CharacterBert-DR+ST & 960h & 95M & 210M & 42.5ms \\
Multimodal DR (Ours) & \textbf{-} & \textbf{-} & \textbf{200M} &  \textbf{21.9ms} \\ \bottomrule
\end{tabular}%
}
\caption{Comparison of our \textit{Multimodal DR} and \textit{ASR-Retriever} pipeline w.r.t needs in annotated speech data, the number
of model parameters, and query time.}
\label{tab:params_data_time}
\end{table}
\subsection{Analysis}
\label{sec:analysis}
Our multimodal dense retriever is ASR-free. Thus, there are no ASR errors that can propagate to the retriever. On the contrary, in the \textit{ASR-Retriever} pipeline, the transcribed question that arrives as input to the retriever will often contain mistranscribed words. Nevertheless, not every word in a question is of equal importance. Let us take as examples the following two transcriptions: ``what channel is young sheldon on'' $\rightarrow$ ``what channel is young shelternon'' and ``who took the first steps on the moon in 1969'' $\rightarrow$ ``he took the first steps on the moon in nineteen sixty nine''. Concerning the former, the corruption can lead the retrieval far from the underline entity ``young sheldon'', while in the latter, the error will have a limited impact on the retrieval. We claim that mistranscribing an important word can hurt retrieval performance more than mistranscribing less important ones. To verify our claim, we explore how the importance of the mistranscribed word impacts the retrieval performance of pipelines and how it compares against our ASR-free multimodal dense retriever (\textbf{RQ2}).

\begin{figure}[!h]
        \centering
        \includegraphics[width=1\linewidth]{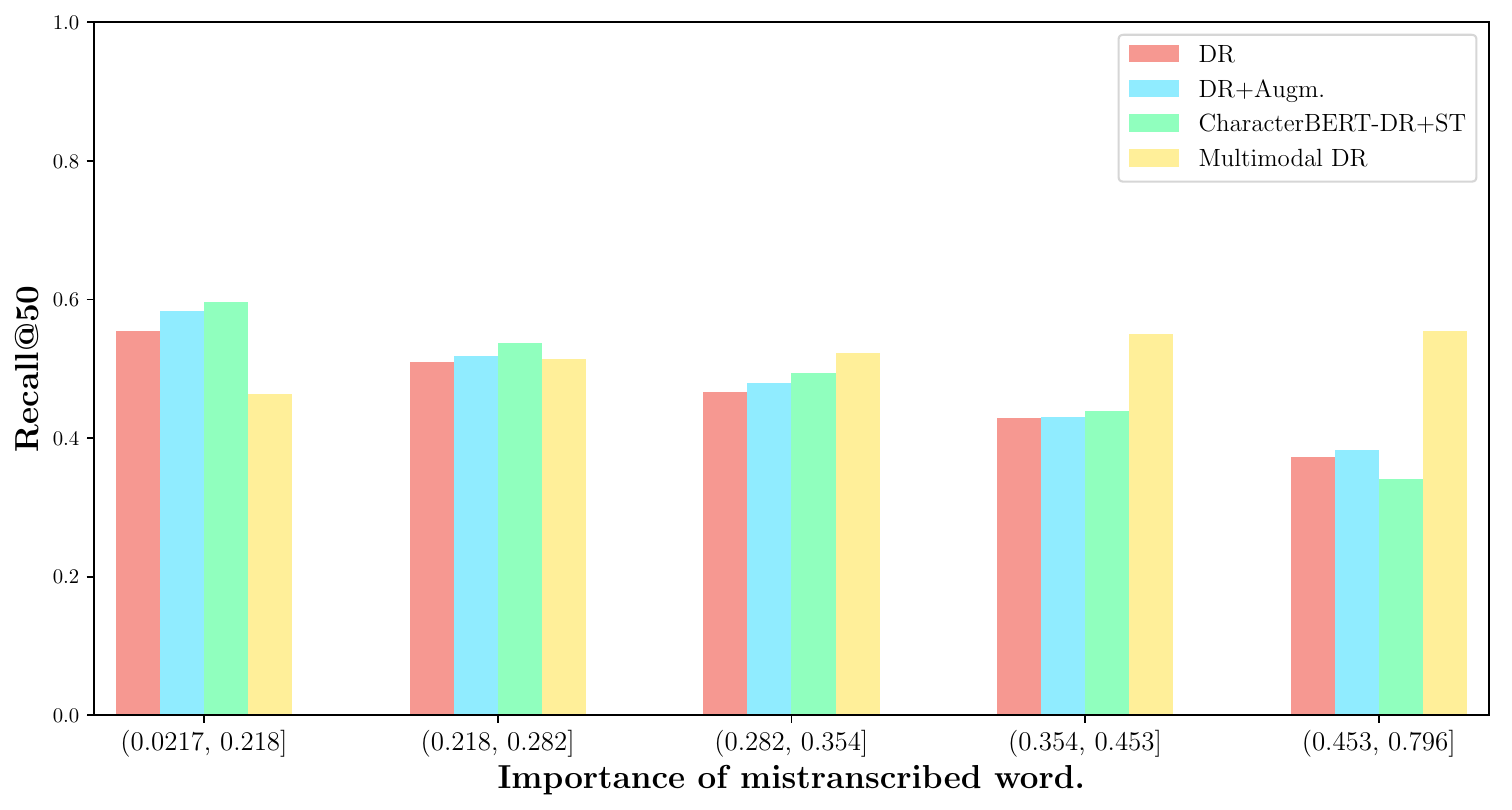}
        \caption{Retrieval results w.r.t the relevant importance of
the mistranscribed words; on Spoken-MSMARCO (dev). For questions with multiple mistranscribed words, we use the word with the highest relevant importance to assign the question to a bin. Bins have ${\sim}1000$ samples.}
        \label{fig:importance}
\end{figure}

For our experiments, we define the relevant importance of a word in a question as the ratio of its IDF to the sum of the IDFs of every word in the question \cite{DBLP:conf/sigir/SidiropoulosK22}. Figure \ref{fig:importance} shows that as the importance of the mistranscribed word increases, there is a dramatic drop in the retrieval performance of the \textit{ASR-Retriever} pipelines. At the same time, as more significant words are corrupted due to the failure of the ASR model, following our ASR-free multimodal dense retriever method is a promising alternative.

%%%
%%%

Our ASR-free multimodal dense retriever is trained on spoken questions. On the other hand, in an \textit{ASR-Retriever} pipeline, the retrieval model is trained on ASR transcriptions. As a result, the retriever encounters ASR mistranscription during training, similar to the ones it encounters during inference. However, a mistranscription of a particular word is strongly connected to its context. For instance, we can have the following two mistranscriptions for the word ``exxon'': ``when did standard oil of new jersey become exxon'' $\rightarrow$ ``when did standard oil of new jersey become exon''
and ``where are exxon's refineries located'' $\rightarrow$ ``where a rexons refineries located''.

\begin{table*}[hbt!]
\resizebox{\textwidth}{!}{%
\begin{tabular}{@{}l|l|c|ccc|ccc@{}}
\toprule
\multirow{2}{*}{} & \multirow{2}{*}{Training Questions} & \multirow{2}{*}{End-to-end} & \multicolumn{3}{c|}{Unseen} & \multicolumn{3}{c}{Seen} \\
                     &      &                 & MRR@10 & R@50 & R@1000 & MRR@10 & R@50 & R@1000 \\ \midrule
DR                   & Transcriptions     &    \xmark & 13.45   & 40.29 & 67.29   & 16.69 & 50.32 & 79.55        \\
DR+Augm.             & Transcriptions \& Typos     &\xmark & 13.41   & 41.08 & 69.63   & 17.18 & 51.75 & 80.83     \\
CharacterBERT-DR+ST  & Transcriptions \& Typos     & \xmark & 15.15  & 40.86 & 67.27   & \textbf{19.75} & \textbf{52.32} & 78.62       \\ \midrule
Multimodal DR (Ours) & Speech     &\cmark & \textbf{18.52}   & \textbf{54.26} & \textbf{82.11} & 15.47  & 50.64 & \textbf{81.58}       \\ \bottomrule
\end{tabular}%
}
\caption{Retrieval performance on Spoken-MSMARCO (dev) for the cases where at inference time (i) the question has a mistranscription that the model encountered during training (Seen), or (ii) has a mistranscription that the model never encountered during training (Unseen). Unseen covers 1,883 samples, while Seen 3,437.}
\label{tab:seen_vs_unseen}
\end{table*}

For \textbf{RQ3}, we study the effectiveness of our ASR-free multimodal dense retriever against \textit{ASR-Retriever} pipelines when the latter encounters previously unseen ASR mistranscribed words. In detail, we explore cases where a particular corrupted word during inference time was not part of the training set. In Table \ref{tab:seen_vs_unseen} we compare the retrieval performance for the cases of (i) previously seen and (ii) previously unseen ASR corrupted words. 

Table \ref{tab:seen_vs_unseen} unveils that a big part of the strong performance we observe on the \textit{ASR-Retriever} pipelines stems from the fact that retrievers are trained on the exact same mistranscriptions they encounter during inference. There is a decrease of more than $10$ points in Recall for all pipelines when they deal with corruption due to ASR that was not part of the training set. Additionally, our multimodal dense retriever significantly outperforms all the pipelines under the unseen scenario.

Inspired by the results in Table \ref{tab:seen_vs_unseen}, we set to study an extreme case of unseen ASR noise. In particular, we study how much the retrieval performance of pipeline approaches deteriorates when their dense retrievers are not explicitly trained on ASR noise (transcriptions). To do so, we train from scratch all the dense retrievers on clean questions instead of transcriptions. We report the results in Table \ref{tab:extreme_unseen}. By comparing the retrieval models when they are explicitly aware of ASR corrupted words during training (as shown in Table \ref{tab:multimodal_vs_pipeline}) vs. when they are not (Table \ref{tab:extreme_unseen}), we see a dramatic drop in performance of more than 20 point drop in Recall. From Table \ref{tab:extreme_unseen}, we can further conclude that for the extreme case where the \textit{ASR-Retriever} pipeline is not trained on ASR noise, following our multimodal dense retriever approach is necessary.

\begin{table}[hbt!]
\resizebox{\columnwidth}{!}{%
\begin{tabular}{@{}llccc@{}}
\toprule
& \begin{tabular}[c]{@{}l@{}}Training \\ questions\end{tabular} & End-to-end & R@1000 & MRR@10 \\ \midrule
BM25                 & -              & \xmark & 45.34  & 6.97   \\
DR                   & Clean          & \xmark & 56.83  & 11.36   \\
DR+Augm.             & Clean \& Typos & \xmark & 64.57  & 13.02  \\
CharacterBERT-DR+ST       & Clean \& Typos & \xmark & 66.94  & 15.75  \\ \midrule
Multimodal DR (Ours) & Speech         & \cmark & \textbf{81.34}  & \textbf{15.77}  \\ \bottomrule
\end{tabular}%
}
\caption{Retrieval results on Spoken-MSMARCO (dev) for the setting where the dense retrievers of \textit{ASR-Retriever} are not explicitly trained on ASR corrupted words.}
\label{tab:extreme_unseen}
\end{table}
\subsection{Ablation Study on Model Training}
\label{sec:model_ablation}
To better understand how different model training schemes affect retrieval performance (\textbf{RQ4}), we perform an ablation on our multimodal dense retriever and discuss our findings below.

\paragraph{Learning Rate}
In traditional dense text retrieval, the same learning rate is used to update all the weights in the retriever \cite{DBLP:conf/emnlp/KarpukhinOMLWEC20,DBLP:conf/naacl/QuDLLRZDWW21}. However, in our multimodal dense retrieval setup, the HuBERT (question encoder) and BERT (passage encoder) models are pre-trained independently to allow the usage of available large-scale unsupervised data. Therefore, there can be disparities between the two modalities that can hurt performance.  To alleviate this problem we follow an alternative setting where the two encoders have different learning rates. In particular, since the language model contains more information than the speech model, we increase the learning rate of the passage encoder by a factor of 10. Comparing the values in the first block of Table \ref{tab:training_strategy}, we find that the choice of learning rate is important for effectively training our multimodal dense retriever.

\begin{table}[h]
\resizebox{\columnwidth}{!}{%
\begin{tabular}{@{}lccc@{}}
\toprule
Pooling & Learning Rate              & AR@20 & AR@100\\ \midrule
first   & p: $2e$-$5$, q: $2e$-$5$ & 50.77 & 64.48\\
first   & p: $2e$-$5$, q: $2e$-$4$ & \textbf{57.25} & \textbf{70.11}\\ \midrule
mean    & p: $2e$-$5$, q: $2e$-$4$ & 56.59 & 69.97\\
max     & p: $2e$-$5$, q: $2e$-$4$ & 53.57 & 67.45\\ \bottomrule
\end{tabular}%
}
\caption{Comparison of different training schemes on Spoken-NQ. We indicate the learning rate of the question and passage encoder as q and p, respectively.}
\label{tab:training_strategy}
\end{table}

\paragraph{Pooling}The next ablation involves different pooling methods for encoding the spoken question into a single vector. Following previous works on dense retrieval, we use the \textit{[CLS]} token embedding output from BERT to encode the text passage. In contrast, this decision is not that straightforward in the case of the spoken question. The HuBERT speech transformer we use for encoding the spoken questions does not have a next-sentence prediction pre-training task as in BERT. Thus, there is no \textit{[CLS]} token available. To this extent, we asses different pooling strategies for encoding the spoken question, namely, max and mean pooling or taking the first embedding of the sequence as a pooling strategy. As we can see in Table \ref{tab:training_strategy}, using the first token embedding output from HuBERT to represent the speech utterance holds the best results.

\paragraph{}For our ablation study we reported results on the test split of Spoken-NQ (Table \ref{tab:training_strategy}). However, we want to clarify that the decision for our best multimodal retriever was based on the development, as discussed in Section \ref{sec:details}.

\section{Related Work}
\label{sec:related_work}
Passage retrieval is a key task in traditional open-domain QA and speech-based open-domain QA. In detail, following the retriever and reader framework, the retriever reduces the search space for effective answer extraction and provides the support context that users can use to verify the answer. Traditional open-domain QA has seen significant advancements with the introduction of dense retrievers \cite{DBLP:conf/emnlp/KarpukhinOMLWEC20,DBLP:conf/naacl/QuDLLRZDWW21, DBLP:journals/corr/abs-2101-00774} that have demonstrated higher effectiveness than bag-of-words methods.

Despite their effectiveness, dense retrievers can still perform poorly in the presence of noise. \citet{DBLP:conf/emnlp/ZhuangZ21} investigated the robustness of dense retrievers against questions with typos. Their work suggested that dense retrievers perform poorly on questions that have typos, and to this extent, they proposed a typo-aware training strategy (data augmentation) to alleviate this problem. \citet{DBLP:conf/sigir/SidiropoulosK22} built upon this and further proposed to combine data augmentation with a contrastive loss to bring closer the representations of a question and its typoed variants. \citet{DBLP:journals/corr/abs-2204-00716} showed that replacing the backbone BERT encoders of the dense retriever with CharacterBERT can increase robustness against typos. They further improved robustness via a novel self-teaching training strategy that distills knowledge from questions without typos into typoed questions.

Recently, \citet{DBLP:conf/cikm/SidiropoulosVK22} studied the robustness of dense retrievers under ASR noise. By employing a pipeline approach with an ASR system and a dense retriever for text retrieval, they showed that dense retrievers, trained only on clean questions, are not robust against ASR noise. They further suggested that training the retriever to be robust against typos can increase the robustness against ASR noise. To the best of our knowledge, this is the first work that studies passage retrieval for speech-based open-domain QA.

On the other end of the spectrum, recent works in speech-based QA explored reading comprehension as a component of the retriever and reader framework. \citet{DBLP:conf/eacl/RavichanderDRMH21} showed that mistranscription from the ASR model leads to a huge performance degradation in the transformer-based reading comprehension models while \citet{DBLP:conf/emnlp/FaisalKAA21} suggested that background differences in the users (e.g., accent) have different impacts on the performance of reading comprehension models.

% spoken qa vs open-domain spoken qa
At this point, we want to highlight the differences between speech-based open-domain QA and spoken QA \cite{DBLP:conf/icassp/YouCZ21, DBLP:conf/interspeech/LeeWLL18} since there can be misconceptions due to the similarity in their names. Spoken QA is a searching through speech task where given a text question and a spoken document the goal is to find the answer from this single spoken document. Therefore, this is a fundamentally different problem compared to the problem we consider in this work.

\section{Conclusions}
In this work, we thoroughly analyzed the behavior of ASR-Retriever pipelines for passage retrieval for speech-based open-domain QA, showing their limitations, and we further introduced the first end-to-end trained, ASR-free multimodal dense retriever in order to tackle these limitations. Our experimental results showed that our multimodal dense retriever is a promising alternative to the \textit{ASR-Retriever} pipelines on shorter questions and under higher word error rate scenarios. Furthermore,  we unveiled that \textit{ASR-Retriever} pipelines show a dramatic performance decrease as the word error rate of the ASR model increases, when important words in the spoken question are mistranscribed, and when dealing with mistranscriptions that have not been encountered during training. To this extent, we showcased that our ASR-free multimodal dense retriever can overcome the aforementioned issues. Finally, despite the limited performance of our proposed method on longer questions, we believe that our thorough analysis can spur community interest in passage retrieval for speech-based open-domain QA.

\section{Limitations}
% mandatory to include section 
In this work, all the models are trained using the hard negatives provided by the original datasets or mined from BM25 \cite{DBLP:conf/emnlp/KarpukhinOMLWEC20} and by employing the base versions of the speech and language transformer models (see Section \ref{sec:details}). Therefore, we leave the application of more complex hard negatives mining techniques, such as mining from the dense index of a previous checkpoint of the dense retriever itself \cite{DBLP:conf/iclr/XiongXLTLBAO21}, and larger models (e.g., BERT-Large and HuBERT-Large) to future work.

As we saw in Section \ref{sec:main_results}, our multimodal dense retriever showed promising results against its \textit{ASR-Retriever} counterparts on shorter questions under high word error rate scenarios. 
We conjecture that the limited performance of our approach on long questions is due to encoding all the information from the speech signal in a single vector, and we leave exploring a multi-vector retrieval approach as future work. \textit{ASR-Retriever} pipelines can produce significantly better results compared to our method for cases where the ASR model can perform high-quality transcriptions with low word error rate. To this extent, we did not experiment with more advanced ASR models, such as the recently proposed Whisper \cite{DBLP:journals/corr/abs-2212-04356}, since our goal in this work was to provide alternatives for cases with high word error rates in the transcriptions.

\bibliography{custom}
\bibliographystyle{acl_natbib}

\end{document}